\documentclass[11pt]{article}
\usepackage[margin=1in]{geometry}
\usepackage[T1]{fontenc}
\usepackage[utf8]{inputenc}
\usepackage{times,amsmath,amssymb,booktabs,array,longtable,graphicx,xcolor}
\usepackage[round,authoryear]{natbib}
\usepackage{microtype,url,enumitem}
\usepackage[colorlinks=true,linkcolor=blue!45!black,citecolor=blue!45!black,urlcolor=blue!45!black]{hyperref}
\usepackage[font=small,labelfont=bf]{caption}
\graphicspath{{figures/}}
\newcommand{\R}{\mathbb{R}}

\title{\bfseries SpikeWorld: Adapting Frozen World Models\\with Compact State}
\author{\normalsize Ziqiao Yu\thanks{Email: \href{mailto:yuziqiao@didiglobal.com}{\texttt{yuziqiao@didiglobal.com}}}, Hongming Zhang, Kai Wang, Runze Wei, Runze Cui,\\
\normalsize Yizhou Tan, Bowen Zhou, Siming Lu, Meijun Guo, Qingfeng Tang}
\date{December 2026}
\hypersetup{pdftitle={SpikeWorld: Adapting Frozen World Models with Compact State},pdfauthor={Ziqiao Yu, Hongming Zhang, Kai Wang, Runze Wei, Runze Cui, Yizhou Tan, Bowen Zhou, Siming Lu, Meijun Guo, Qingfeng Tang}}
\begin{document}
\raggedbottom
\maketitle
\begin{abstract}
World models can learn from their own prediction errors after deployment, but
updating a shared model also changes the representation used by its other tasks.
We introduce SpikeWorld, a predictive--semantic spiking model that keeps its
trained parameters fixed and learns deployment corrections in compact external
state. We train a shared causal core to predict future sensory features and
action outcomes, together with semantic recognition and cross-modal association.
At deployment, we use observed prediction errors to update residual matrices
that refine future predictions and candidate scores that select bounded action
corrections. On a six-stage stream of registered actuator changes,
28.8 KB of mutable numeric state reduces prediction error by 19.3\%, compared
with 21.9\% for the evaluated prediction-path fine-tuning configuration using
1.93 MB. We test the components separately to identify which improves prediction and
which changes actions. A closed-loop study shows that the action correction
improves reward. Throughout adaptation, the trained model and its semantic
outputs on fixed inputs remain unchanged. These results show that a predictive model can use deployment experience to
improve its predictions while preserving its learned semantic mapping.

\end{abstract}
\section{Introduction}
Imagine a robot reaching for an object after its motors have become weaker.
The same command now produces a different movement. A world model can detect
this discrepancy: it predicts the outcome of a command, observes the actual
outcome, and measures the difference. The question is how to use that new
experience. Updating the model is one option. Another is to keep the trained
model intact and learn a small correction alongside it.

This second option is useful when a representation serves several purposes.
A model may predict future observations, recognize semantic categories, and
associate information across sensors. Deployment feedback about an actuator
provides a direct signal for dynamics prediction, while the model's other
outputs may need to remain stable. A separate correction can learn from this feedback while the trained
representation remains fixed.

World models learn representations through prediction
\citep{ha2018worldmodels,hafner2019planet,hafner2020dreamer}.
For a sequence predictor, the timing and order of observations can matter as
much as their content. We choose a spiking temporal core to make recent input
history part of the computation: membrane variables accumulate incoming signals,
leak over time, and emit spikes when their accumulated value crosses a threshold.
This gives the model a mechanism for representing temporal patterns, which
surrogate-gradient training can learn to use \citep{yu2026beyondrate}.
Causal attention combines these signals using only the observed prefix.
Our controlled event-model probes support this temporal motivation: future
prediction is more accurate with ordered sequences than when event features
are shuffled or sorted onto a regular time grid
(Section~\ref{sec:temporal_evidence}).

Audio, visual features, rendered words, and state--action inputs enter through
appropriate adapters. Each adapter preserves its stream's prediction target; all streams use the
same temporal core. We use \emph{next-spike prediction} to describe this
event-oriented learning approach and specify the actual next-feature and
action-outcome targets in Section~\ref{sec:model}.

\begin{figure}[tbp]
\centering\includegraphics[width=\linewidth]{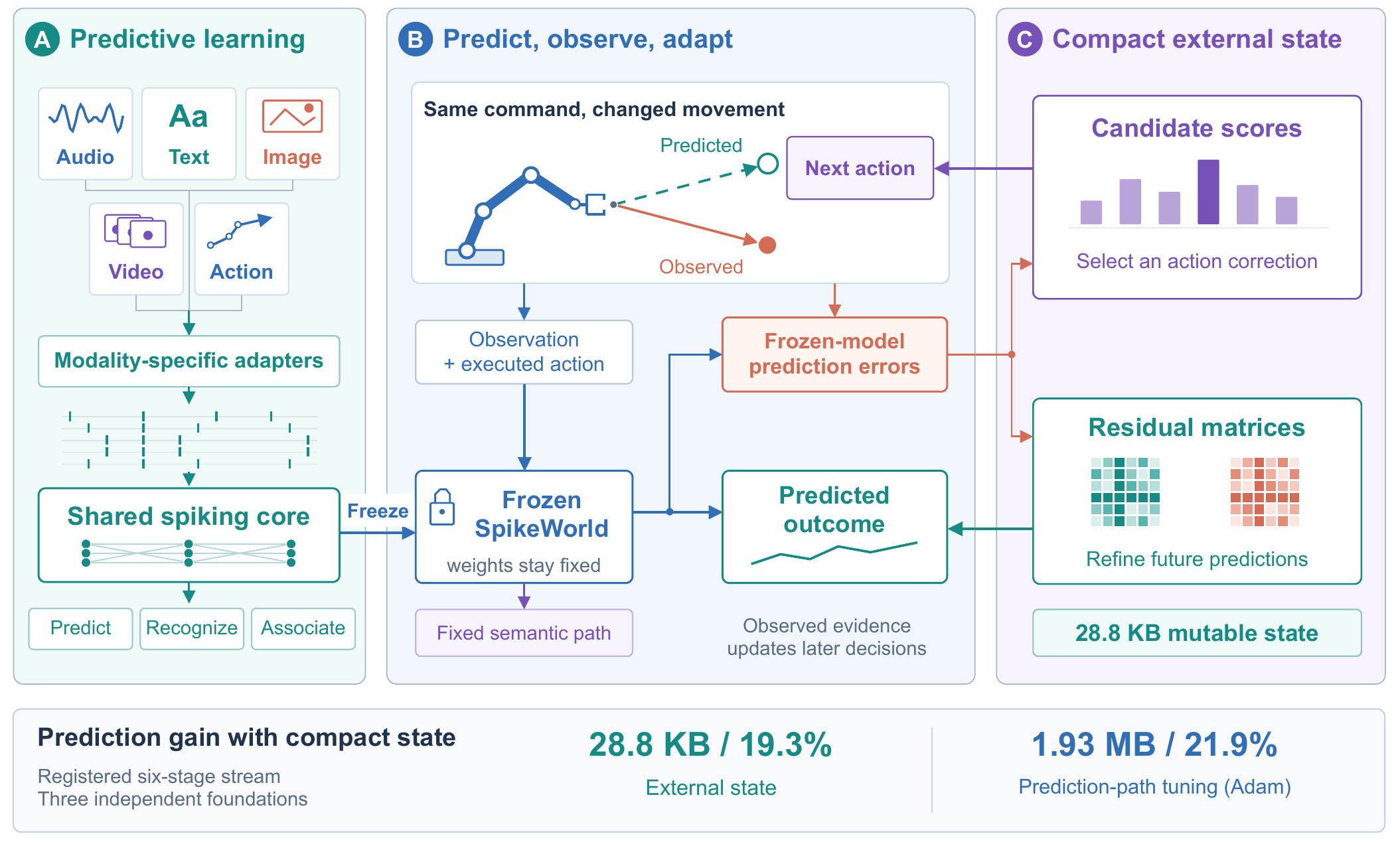}
\caption{\textbf{A shared predictive model, with room to learn outside its weights.}
\textbf{A}, sensory and action inputs enter through modality-specific adapters
and train a shared spiking core for prediction, recognition, and association.
\textbf{B}, the selected checkpoint is frozen. The robot illustration shows
how changed actuation creates a difference between expected and observed movement.
Outcomes evaluated through the frozen predictor supply online learning evidence.
\textbf{C}, candidate scores select a bounded correction to a later command
(purple); residual matrices refine the prediction (teal). The semantic path
retains its fixed-input mapping. The bottom strip summarizes the main-stream
prediction--storage comparison. Icons, trajectories, scores, and matrix patterns
are schematic; the bottom-strip values are measured results.}
\label{fig:workflow}
\end{figure}

Figure~\ref{fig:workflow} connects offline learning to deployment.
The frozen predictor supplies an error signal when a new outcome is observed.
Residual matrices use this signal to refine predictions; separate candidate
scores select a bounded command correction from registered actuator changes.
An existing control policy proposes each command. The interface adjusts it,
and we disable each external component in turn to identify its effect.

On a six-stage stream of source, attenuation, and shear conditions, the
external state recovers most of the prediction improvement of the evaluated
prediction-path fine-tuning configuration with much less mutable storage.
Component interventions identify the matrices' prediction contribution, and
a separate closed-loop experiment connects command correction to reward.

We make three contributions. We train and evaluate a shared predictive--semantic
spiking model with sensory and action targets. We design an online interface
that stores corrections in bounded external state and preserves the trained
semantic mapping. We measure its prediction--storage trade-off and test the
separate roles of its state components. The appendix provides the full equations,
protocols, statistical methods, and additional comparisons.
Code, editable figures, and recorded experimental outputs are available at
\url{https://github.com/Oooorca/SpikeWorld}.

\section{A shared model for prediction and semantics}
\label{sec:model}
We train the model to predict future observations and recognize categories
from a shared feature sequence. Figure~\ref{fig:targets} shows the training
tasks. Panel A predicts the next sensory slice from the observed prefix.
Panel B predicts the later outcome of an action. Panel C shows how membrane
accumulation and spikes process sequences, and which losses train the readouts.

\subsection{Inputs, features, and spikes}
Let $m$ identify a sensory stream and let
$x^m_{1:T}\in\R^{T\times D_m}$ contain $T$ feature slices with $D_m$ channels
per slice. A slice is the numerical feature vector presented at one sequence
position. The implementation uses $T=16$, with modality-dependent channel
counts. Learned input adapters and modality embeddings map these features
into a shared width-$d$ representation, where $d=96$.
The audio, image, text, and video adapters and their data sources are described
in Appendix~\ref{app:training}.

\begin{figure}[tbp]
\centering\includegraphics[width=\linewidth]{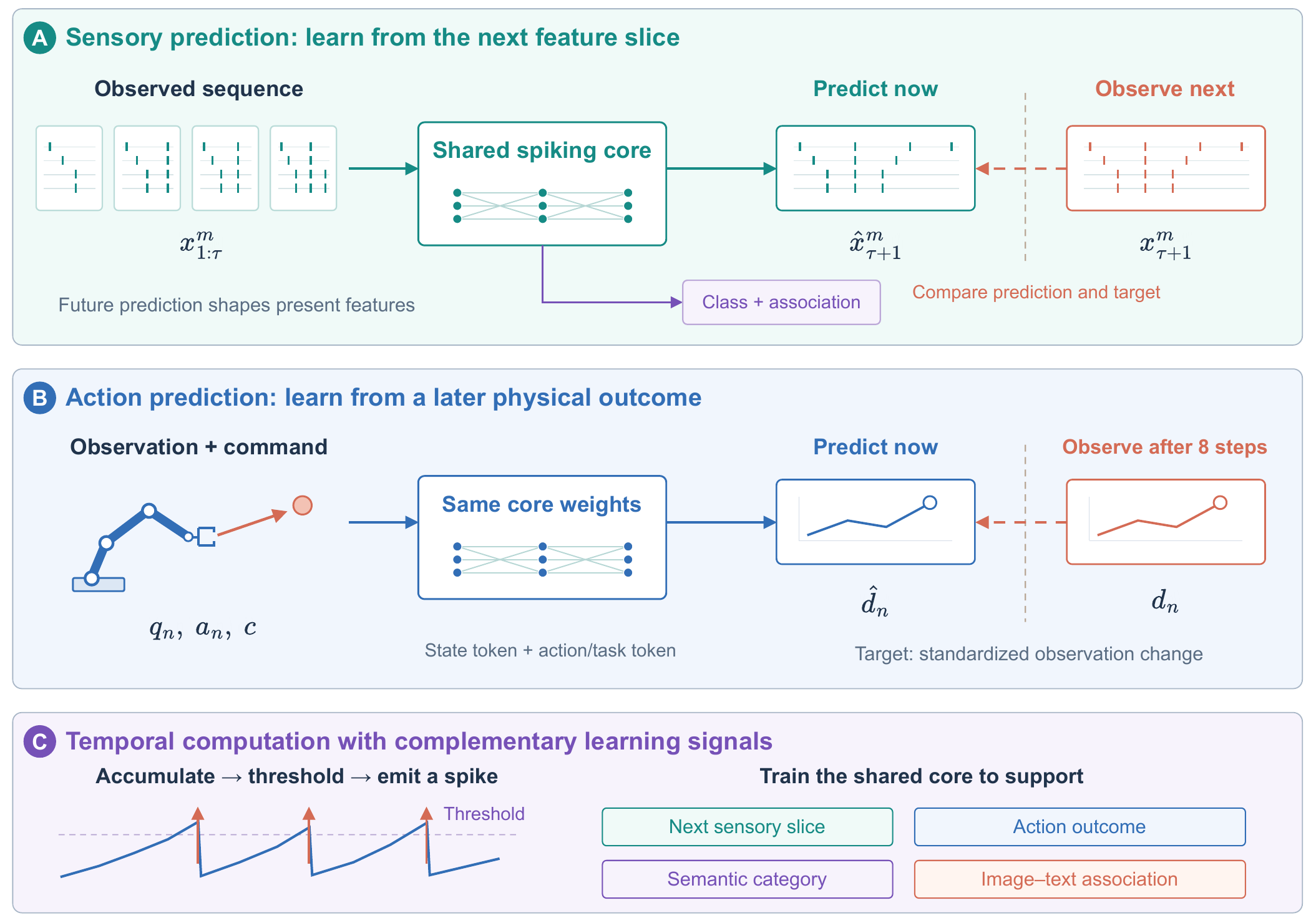}
\caption{\textbf{Predict the next sensory slice and the later consequence of an action.}
\textbf{A}, the observed sensory prefix produces a next-slice prediction; the
next observation supplies its target. The same features also support class
recognition and association. \textbf{B}, current observation, command, and task
enter the same core to predict a standardized observation change, whose target
becomes available after eight environment steps. Dashed vertical lines separate
prediction time from target availability; coral arrows indicate target comparison.
\textbf{C}, membrane accumulation and thresholded spikes provide temporal
computation, while four objectives train complementary readouts of the shared
core. Sensory position $\tau$ and action decision $n$ use different clocks.
Feature patterns, trajectories, and membrane traces are schematic illustrations.}
\label{fig:targets}
\end{figure}

A spiking neuron retains a membrane variable between sequence positions.
Figure~\ref{fig:targets}C illustrates how incoming information accumulates
until the membrane crosses a threshold and emits a spike.
For sensory position $\tau$, its update is
\begin{equation}
 v_\tau=\beta u_{\tau-1}+I_\tau,\qquad
 s_\tau=\mathbf{1}[v_\tau>V_{\mathrm{th}}],\qquad
 u_\tau=v_\tau-V_{\mathrm{th}}s_\tau.
 \label{eq:lif}
\end{equation}
Here $I_\tau$ is the incoming current, $u_{\tau-1}$ is the previous post-reset
membrane value, $v_\tau$ is the value before reset, and $s_\tau$ is a binary
spike. The indicator $\mathbf{1}[\cdot]$ is one when its condition holds and
zero otherwise. The decay $\beta=0.85$ retains part of the preceding value, and the
threshold $V_{\mathrm{th}}=1$ determines whether a spike is emitted. The last
operation subtracts the threshold after a spike. During offline training,
surrogate derivatives supply a gradient through the threshold
\citep{neftci2019surrogate,yu2026beyondrate}.

Equation~\eqref{eq:lif} explains the temporal role of the spiking model.
An input can affect later membrane values even if it does not immediately
produce a spike. Decay gives more recent inputs greater weight, so rearranging
the same inputs can change when the threshold is crossed. The next-slice
objective trains the core to use this history to predict what follows.
Membrane accumulation supplies local temporal context; causal attention
combines information across the available sequence positions.

The shared core has two Transformer blocks and four attention heads. Each
query attends to at most four causal positions, including itself. Spikes enter
the attention and feed-forward paths, while residual connections carry numerical
features. The learned representation is a numerical feature sequence.
This implementation evaluates finite input
sequences on conventional hardware; its sparse attention is specified in
Appendix~\ref{app:training}.

\subsection{Predicting the next sensory slice}
In Figure~\ref{fig:targets}A, the modality-specific output head uses positions
$1$ through $\tau$ to produce
$\hat x^m_{\tau+1}$, a prediction of the next feature slice. The hat marks a
prediction; $x^m_{\tau+1}$ without the hat is the observed target.
The prediction loss for stream $m$ is
\begin{equation}
 \ell_m=\frac{1}{(T-1)D_m\,p_m}
 \sum_{\tau=1}^{T-1}\|\hat x^m_{\tau+1}-x^m_{\tau+1}\|_2^2.
 \label{eq:next}
\end{equation}
The squared norm $\|\cdot\|_2^2$ sums squared errors across feature channels.
The outer sum then adds these errors over the available next-slice pairs.
Dividing by $(T-1)D_m$ averages over positions and channels. The positive
constant $p_m$ is the training-set error of the persistence baseline, which
predicts the next slice by copying the current one. This normalization puts
modalities with different numerical scales on comparable footing. The
sensory loss $\mathcal L_{\mathrm{next}}$ gives equal weight to audio, text,
image, and video groups. The audio group first averages its two sources,
SHD and SSC; Appendix~\ref{app:losses} gives the combined expression.

The core processes spikes, while the output adapters predict encoded feature
slices in each stream's feature space. The targets can therefore contain
non-binary values. Here, next-spike refers to causal prediction over these
event-oriented sensory sequences.

\subsection{Predicting action outcomes}
Figure~\ref{fig:targets}B uses a different horizon: the consequence of executing
a command. Use $n$ for an action decision and $t_n$ for its starting environment step.
An observation $q_n=q(t_n)\in\R^{D_q}$ has $D_q=39$ coordinates and contains
the simulator's numerical observation at the start of that decision. A command
$a_n\in[-1,1]^4$ and task identifier $c$ are supplied to the action branch.
Let $q_n^+=q(t_n+H)$ be the observed outcome after horizon $H=8$ environment
steps. The superscript $+$ marks this later observation. We predict how the
observation changes over the interval, using the standardized target
\begin{equation}
 d_n=(q_n^+-q_n-\mu_\Delta)\oslash\sigma_\Delta,
 \qquad
 \mathcal L_{\mathrm{act}}=\frac{1}{D_q}
 \|g_\theta(q_n,a_n,c)-d_n\|_2^2.
 \label{eq:action}
\end{equation}
Here $g_\theta$ is the action-conditioned predictor with trained parameters
$\theta$. The vectors $\mu_\Delta,\sigma_\Delta\in\R^{D_q}$ are the mean and
scale of training observation changes; $\oslash$ is coordinate-wise division.
Thus $d_n$ is a vector of observed changes, distinct from the scalar hidden
width $d$ introduced above.
The factor $1/D_q$ averages error over output coordinates. In the implementation
it is $1/39$, a dimensional normalization rather than a tuned coefficient.
The state and action/task adapters provide two tokens to the shared core;
the frozen linear endpoint head maps the final action feature to $D_q$ outputs.

\subsection{Learning predictions and semantic readouts together}
The four losses in Figure~\ref{fig:targets}C train prediction, recognition,
and association from the same core. The coupled semantic heads read its
features to predict category logits, the class scores before conversion to
probabilities.
Their cross-entropy loss is $\mathcal L_{\mathrm{sem}}$. A binding head maps
image and rendered-text features to normalized vectors, and a symmetric
contrastive loss $\mathcal L_{\mathrm{bind}}$ associates examples with the same
digit identity. Joint training minimizes
\begin{equation}
 \mathcal L_{\mathrm{joint}}=
 \mathcal L_{\mathrm{next}}+\mathcal L_{\mathrm{sem}}
 +\lambda_b\mathcal L_{\mathrm{bind}}+\lambda_a\mathcal L_{\mathrm{act}},
 \qquad \lambda_b=0.25,\quad\lambda_a=4.
 \label{eq:joint}
\end{equation}
The fixed coefficients $\lambda_b$ and $\lambda_a$ set the relative weights
of association and action prediction in the training loss.
Equation~\eqref{eq:joint} trains the core to preserve
information useful for both future prediction and present classification.
Appendix~\ref{app:losses} expands the classification and contrastive terms.

We evaluate two training stages separately.
The \emph{sequential checkpoint} begins with sensory pretraining and then trains
the action path for 800 updates while holding the core fixed. This is the
checkpoint used in the main deployment comparison. The \emph{joint checkpoint}
adds 200 shared-core updates with Equation~\eqref{eq:joint}. A separate study
compares these stages to measure the effect of joint learning. Once a checkpoint
is selected for deployment, all of its trained parameters are frozen.

\section{Learning corrections outside the frozen model}
\label{sec:adapt}
Consider again the weakened motor. We compare the model's predicted outcome
with the outcome observed in the environment, then use the difference to
update two small objects. A score vector identifies a suitable command
correction; a residual matrix improves the predicted outcome. The following description uses
the decision index $n$ and target $d_n$ defined in Equation~\eqref{eq:action}.

\subsection{Predict first, update after observation}
Let $\theta_0$ denote the predictor parameters from the checkpoint selected for
deployment. These parameters, the semantic heads, the mechanism router, and
the underlying control policy are frozen. The router is an offline-trained
classifier that chooses among source (unchanged actuation), shear, attenuation,
and noise conditions. The online interface
receives the current observation, executed command, task identifier, and eventual
outcome. Reward, success, and the true actuator change are evaluation variables,
not online update inputs.

Every prediction is recorded before its target is observed. After the horizon
has elapsed, the system computes an error and updates external state for later
decisions. The first four transitions in an episode supply evidence to the
frozen mechanism router. The route locks after the fourth outcome arrives. The system can then update
the persistent state for the first time. Consequently, decision five is
the earliest command that can change. Appendix~\ref{app:algorithm} gives the
complete schedule, including the different lifetimes of episode prefixes and
persistent matrices.

\subsection{Scores select an actuator correction}
The model compares a bank of $J=32$ transformations: identity, shear, and
attenuation candidates. Identity leaves the command unchanged. Shear couples
the first two action coordinates, so that an $x$ command also affects $y$.
Attenuation reduces the realized movement in the $x/y$ coordinates. The bank's
values are registered offline; the online system infers a candidate from
observed transitions.

For candidate $j$, let $T_j$ be its command transformation. Once outcome $d_n$
is available, the candidate receives loss
\begin{equation}
 \ell_{n,j}=\frac{1}{D_q}
 \|g_{\theta_0}(q_n,T_j(a_n),c)-d_n\|_2^2,
 \qquad S_j\leftarrow S_j+\ell_{n,j}.
 \label{eq:scores}
\end{equation}
The cumulative score $S_j$ starts at zero and adds admitted transition losses
$\ell_{n,j}$; lower scores better explain observed outcomes. Within the routed
family $\rho$, the smallest score selects the estimated transformation.
Scoring queries the original frozen predictor, without residual matrices.
Source and noise routes keep commands unchanged and skip matrix updates.

Let $a_n^*$ be the command proposed by the underlying policy, and let
$\tilde a_n$ be the selected inverse transformation applied to that command.
The controller executes
\begin{equation}
 a_n=a_n^*+\operatorname{clip}(\tilde a_n-a_n^*,-\delta,\delta),
 \qquad \delta=0.05.
 \label{eq:command}
\end{equation}
The star marks the policy's desired command; the tilde marks the inverse's
proposed command. The final command $a_n$ limits how far the proposal can move
from the desired command. The clip acts separately on each coordinate, with
$\delta$ setting the maximum change in either direction.
If this capped command would lie outside the allowed action range, the system
uses $a_n^*$ instead. The inverse maps, gain floor, and source/noise behavior
are given in Appendix~\ref{app:algorithm}.

\subsection{Matrices refine the prediction}
After selecting a candidate, the frozen core computes hidden feature
$h_n\in\R^d$ using the current observation, task, and transformed command.
This is an internal feature vector computed for the current input.
We retain the fixed offline mean
$\bar h\in\R^d$ and projection $B\in\R^{r\times d}$, where $r=30$.
The projection is obtained by principal component analysis of offline action
features. Each correctable mechanism family has its own matrix
$C_\rho\in\R^{d\times r}$, initialized to zero. Prediction becomes
\begin{equation}
 z_n=B(h_n-\bar h),\qquad
 \tilde h_n=h_n+C_\rho z_n,\qquad
 \hat d_n=W_{\mathrm{out}}\tilde h_n+b_{\mathrm{out}}.
 \label{eq:correction}
\end{equation}
The compressed feature $z_n\in\R^r$ retains $r$ directions; $C_\rho$ expands
it into a learned correction. The tilde marks the corrected hidden feature,
and $\hat d_n$ predicts the observation change. The readout weight
$W_{\mathrm{out}}\in\R^{D_q\times d}$ and bias $b_{\mathrm{out}}\in\R^{D_q}$
remain frozen. Features $h_n,z_n$ are computed anew for each input;
$C_\rho$ retains what was learned across observed transitions.

After observing an outcome and updating the candidate scores, the active
matrix takes a normalized gradient step on prediction mean squared error
(MSE). The step length is $\eta=0.02$. After each step, the matrix is scaled back
if its Frobenius norm, the square root of the sum of squared entries, exceeds
$R_C=0.20$. Appendix~\ref{app:matrix-update} gives the exact update.
Candidate scoring and command correction do not read these matrices, so
removing them isolates their prediction effect without changing actions.

\subsection{What remains fixed, and what occupies memory?}
The trained semantic computation has no read connection to the external
matrices or scores. Given the same sensory input and inference settings,
it therefore returns the same logits (pre-softmax class scores) before and
after online adaptation.
We verify this with parameter and fixed-input-output hashes. This property
preserves the learned mapping; semantic outputs can still differ when the
robot observes different inputs.

The online allocation contains two residual matrices, the score vector and
count, compatibility padding, and a four-transition routing prefix. Their peak
total is 28,848 bytes of mutable numeric state. This is the memory that can
change during adaptation, separate from the frozen model and transient
inference activations. Table~\ref{tab:ledger} gives the exact arithmetic.
The bounded prefix is cleared after routing, and the matrices and scores retain
experience across the stream.

\section{Evaluation design}
\label{sec:evaluation}
We test whether the shared core learns predictive and semantic features,
how much mutable state online adaptation requires, and which component changes
behavior. The experimental units and statistical methods are summarized in
Appendix~\ref{app:protocols}.

\paragraph{Representation and joint learning.}
The sensory sources include SHD/SSC audio \citep{cramer2022heidelberg}, retinal
digit images and rendered words, and a causal video-feature cache. Action
sequences use Meta-World reach, push, pick-place, and door-open tasks
\citep{yu2019metaworld}. We call a sensory-pretrained checkpoint a
\emph{foundation}; an action seed specifies a fresh initialization of its
action branch. Five action initializations sharing one sensory
foundation compare sequential training, action-only updating, and joint
training. Three additional controlled event-model studies examine temporal
structure, prediction of future observation statistics, and image/audio factor
association. Each uses three seeds and probe models trained separately from the deployment
checkpoints.

\paragraph{The main deployment stream.}
Three independently trained foundations, each with three action seeds, produce
nine paired runs. Each uses the sequential checkpoint. All methods see the
same six-stage layout: source, attenuation, source, shear, attenuation, and
source return. Every stage includes the four manipulation tasks. The tested
shift values belong to the registered bank. Shifted trajectories come from an
execution archive separate from router calibration. Source stages revisit
calibration episodes, so source-return measurements quantify calibrated reentry.

The five configurations are frozen prediction, external state, action-adapter
tuning, prediction-path tuning, and prediction-path tuning with replay. All
three trainable comparison arms use Adam. ``Prediction-path tuning'' refers to
the deployed prediction path, not every parameter in the multimodal model.
We count mutable parameters, persistent optimizer tensors, state and routing
buffers, and replay where applicable. We also record update latency using
CUDA-synchronized measurements after warm-up.

The main study replays a common recorded trajectory in causal order: each
prediction is evaluated before its outcome is used to update the method.
Each run contains 348 transitions, of which 174 are shifted. The primary
score uses the 126 shifted transitions remaining after excluding the first
four transitions of every episode, which establish its route.
For method $M$, prediction gain compares its mean error $E_M$ with the frozen
predictor's error $E_{\mathrm{fr}}$ on the same evaluated transitions:
\begin{equation}
 G_M=100\left(1-\frac{E_M}{E_{\mathrm{fr}}}\right)\%.
 \label{eq:gain}
\end{equation}
A positive gain means smaller error. Action tracking applies the known actuator
map to each corrected command and measures its distance from the desired
command. Replay gives every method the same observations for the prediction
comparison. In the closed-loop test below, corrected commands change subsequent
observations, and reward measures the resulting behavior. Source-return correction measures
the size of unnecessary command adjustments upon calibrated reentry. Action
seeds are averaged within each foundation. Confidence intervals (CIs) use
20,000 bootstrap draws over the three foundation means, preserving paired
method comparisons.

\paragraph{Component and closed-loop tests.}
The component study compares full state, score-only state, and removal of the
inverse. A separate 36-trace check verifies that bounding the routing history
preserves the original implementation's observable behavior. Closed-loop
attenuation uses five model seeds, three tasks, and five environment replicas:
75 trajectories per method. The tasks are reach, push, and pick-place, with
attenuation gain 0.4. Each trajectory contains 13 decisions over 104 environment
steps. Adaptation state carries across the five replicas within each model/task
pair; a new controller is created for the next pair. Reward and success are
evaluation outputs, never adaptation inputs. Its CIs use paired resampling
of model, task, and replica axes. The live-control tuning and replay controllers
are defined separately in Appendix~\ref{app:control}.

\paragraph{Classical calibration and training-stage diagnostics.}
A matched six-stage replay compares the external state with routed actuator-space
recursive least squares (RLS), using a two-by-two estimate for action coupling.
The closed-loop study includes a separate scalar RLS reference for attenuation.
Appendix~\ref{app:additional} also reports a one-foundation development study
of direct output-space calibration, including normalized least mean squares
(NLMS), RLS, and a low-rank readout update. We report each comparison under its own protocol because the calibrators
correct different quantities.
A further one-foundation pilot applies the same external-state interface to
different offline training stages.

\section{Results}
\label{sec:results}
\subsection{A compact prediction--storage trade-off}
\begin{figure}[tbp]
\centering\includegraphics[width=\linewidth]{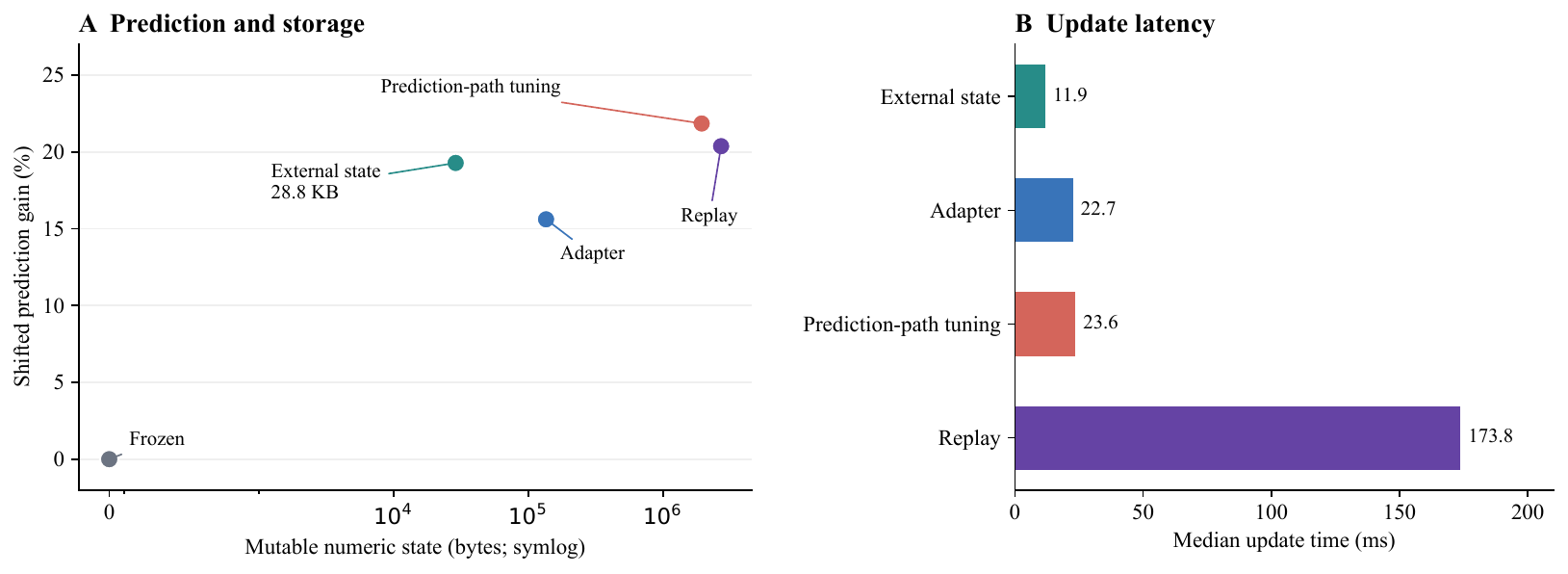}
\caption{\textbf{External state provides a compact deployment configuration.}
Left: prediction gain against peak mutable numeric storage for the five main
arms. Labels identify every point; the horizontal axis uses a symmetric-log
scale to include frozen prediction at zero bytes. Right: median update latency
for the adaptive arms. All arms use the same three-foundation, three-action-seed
design. These are configuration-specific memory and latency measurements.}
\label{fig:tradeoff}
\end{figure}

External state improves shifted prediction MSE by 19.28\% with 28,848 mutable
bytes. Prediction-path tuning reaches 21.85\% with 1,932,488 bytes under the
evaluated Adam configuration. The difference in gain is 2.57 percentage
points; the difference in mutable storage is about 67-fold.
Figure~\ref{fig:tradeoff} and Table~\ref{tab:main} place this result beside the
other evaluated configurations. The external-state point is non-dominated
among these five configurations: no other point has both less mutable storage
and greater mean prediction gain.

\begin{table}[tbp]
\centering\small
\caption{Main six-stage replay comparison. Gains use 126 shifted transitions per run after routing prefixes; all trainable arms use Adam. Mutable state excludes frozen weights and transient activations.}
\label{tab:main}
\begin{tabular}{@{}lrrrr@{}}
\toprule
Method & Pred. gain (\%) & Tracking (\%) & Bytes & Update (ms) \\
\midrule
Frozen & 0.00 & 0.00 & 0 & -- \\
External state & 19.28 & 3.80 & 28,848 & 11.90 \\
Adapter (Adam) & 15.61 & 3.47 & 135,232 & 22.72 \\
Prediction-path tuning (Adam) & 21.85 & 2.91 & 1,932,488 & 23.62 \\
Replay (Adam) & 20.38 & 3.35 & 2,693,128 & 173.83 \\
\bottomrule
\end{tabular}
\end{table}

The external-state mean prediction gain has a foundation-bootstrap 95\% CI
of 15.71\%--21.80\%. Its mean tracking gain is 3.80\%, and calibrated-source
return produces zero command correction. Median update time is 11.9 ms,
approximately half that of the adapter and prediction-path tuning arms.
External state thus improves prediction with a small allocation and a short
update time in this comparison. Appendix~\ref{app:extended} reports the
values for each foundation.

In the separate matched actuator-identification replay, external state obtains
19.72\% prediction gain and the routed matrix RLS obtains 0.58\%. That
comparison isolates actuator-space identification on the six-stage layout.
In the separate development protocol in Appendix~\ref{app:additional}, direct
output-space calibrators achieve larger prediction gains with less state.
Where the correction is applied therefore matters to this comparison.

\subsection{Prediction and action gains come from different components}
Table~\ref{tab:roles} makes the role of the residual matrices explicit.
Relative to scores alone, matrices add 9.22 percentage points of prediction
gain under shear and 25.12 points under attenuation. Yet the full and
score-only variants execute exactly the same commands in the paired
intervention. Removing the inverse makes action improvement zero.
The matrices improve predictions; the scores and inverse change commands.
The interventions separate these two effects of online adaptation.
\begin{table}[tbp]
\centering\small
\caption{Component intervention. Full and score-only commands are bitwise equal; matrix contribution is their difference in prediction gain.}
\label{tab:roles}
\begin{tabular}{@{}lrrrr@{}}
\toprule
Shift & Full gain (\%) & Scores (\%) & Matrix (points) & Tracking (\%) \\
\midrule
Shear & 8.13 & -1.09 & 9.22 & 43.99 \\
Attenuation & 32.63 & 7.51 & 25.12 & 3.88 \\
\bottomrule
\end{tabular}
\end{table}

The bounded-memory implementation preserves this behavior. Across the
36 traces in nine matched four-controller scenarios, the original and bounded
versions agree in observable
fields, and the first changed command occurs at decision five. This verifies
that the small runtime footprint does not depend on retaining an expanding
transition history. Throughout the main deployment runs, parameter hashes
and fixed-input semantic-logit hashes remain unchanged.

\subsection{The action correction improves closed-loop reward}
\begin{figure}[tbp]
\centering\includegraphics[width=0.95\linewidth]{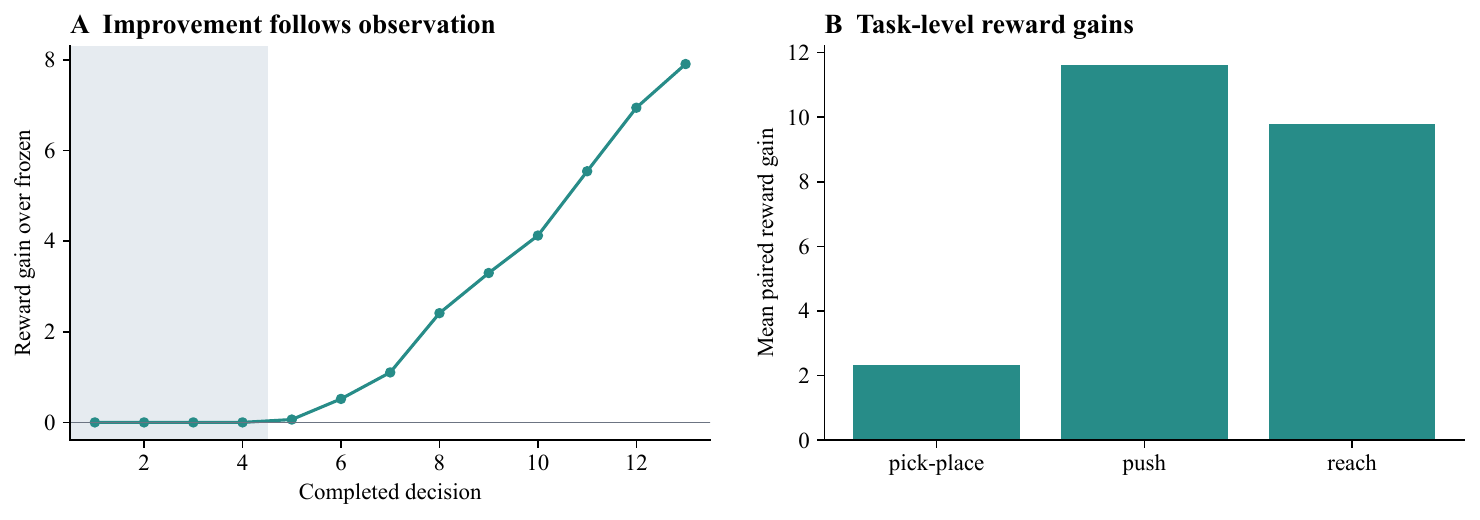}
\caption{\textbf{Behavior changes after evidence is available.}
Left: mean cumulative reward difference between external state and frozen
control, plotted at completed decisions. The shaded prefix contains the first
four decisions; decision five is the earliest corrected command. Right: paired
reward gains by task. The separate attenuation study uses 75 trajectories per
arm. Reward is used for evaluation, not for updating external state.}
\label{fig:control}
\end{figure}

In registered attenuation, external state raises mean cumulative reward from
414.68 to 422.58. The paired reward improvement is 7.90, with a 95\% CI from
2.48 to 14.06. Figure~\ref{fig:control} shows how the reward difference
develops after the observation prefix; Table~\ref{tab:control} includes all
closed-loop comparison arms. The reward gain is positive on each of the three
tasks and across all five model seeds. These results show that the score-and-inverse path improves behavior in this
attenuation study, alongside the matrices' separate effect on prediction.

\begin{table}[tbp]
\centering\small
\caption{Closed-loop attenuation: 75 trajectories per arm; adaptation carries across replicas within each model/task pair. Reward and success are evaluation-only. Controller definitions appear in Appendix~\ref{app:control}.}
\label{tab:control}
\begin{tabular}{@{}lrr@{}}
\toprule
Method & Mean reward & Success (\%) \\
\midrule
Frozen & 414.68 & 53.33 \\
Prediction-model tuning (AdamW) & 422.19 & 66.67 \\
Oracle inverse (uncapped correction) & 485.50 & 80.00 \\
Gain-memory replay & 428.59 & 60.00 \\
Scalar RLS & 434.41 & 65.33 \\
External state & 422.58 & 66.67 \\
\bottomrule
\end{tabular}
\end{table}

\subsection{Shared predictive training supports several readouts}
Relative to the sequential checkpoint, joint training reduces action MSE by
17.10\% and normalized sensory prediction error by 0.75\%. Semantic accuracy
and image--text class retrieval rise by 1.80 and 1.81 percentage points,
respectively. These paired means use five action initializations
sharing a sensory foundation. Table~\ref{tab:joint} reports their intervals.
These changes show that joint updates improve both predictive and semantic
readouts relative to the sequential checkpoint. The complete three-stage comparison
in Appendix~\ref{app:joint} also shows the trade-off with action-only updates:
joint training improves class retrieval, while action-only updating achieves
the lower action error.
\begin{table}[tbp]
\centering\small
\caption{Joint training relative to the sequential checkpoint. Five action initializations share one sensory foundation. Intervals are paired 95\% bootstrap intervals.}
\label{tab:joint}
\begin{tabular}{@{}lrr@{}}
\toprule
Metric & Mean improvement & 95\% CI \\
\midrule
Action MSE reduction & 17.10\% & [14.04, 20.04]\% \\
Sensory MSE reduction & 0.75\% & [0.68, 0.83]\% \\
Semantic accuracy change & 1.80 points & [1.22, 2.44] points \\
Image--text retrieval change & 1.81 points & [0.19, 3.44] points \\
\bottomrule
\end{tabular}
\end{table}

\paragraph{Temporal structure supports future prediction.}
\label{sec:temporal_evidence}
In a separate controlled event-model study, ordered sequences yield a mean
negative log-likelihood (NLL; lower is better) of 4.91, compared with 6.93 for
the rate-grid control and 11.04 for shuffled order, across three seeds.
The rate-grid control sorts event features and assigns evenly spaced timestamps;
the order control shuffles non-time event features across fixed timestamps.
This comparison shows that temporal organization carries useful predictive
information in the probe task, supporting the motivation for a core that
processes input history. Table~\ref{tab:probes} reports the results; Appendix~\ref{app:probes} gives
the probe protocol. These probe encoders are trained separately from the
deployment checkpoints.

Two further probes test future observation statistics and image/audio factor
associations. The training-stage pilot in
Appendix~\ref{app:bridge} then connects different trained checkpoints to the
same external-state interface. The probes test what the representation learns; the pilot tests how the
external correction behaves on checkpoints from different training stages.

\begin{table}[tbp]
\centering\small
\caption{Controlled event-model probes, three-seed means. Lower NLL/loss and higher accuracy are better. The studies test temporal structure, future statistics, and factor association with separate models.}
\label{tab:probes}
\begin{tabular}{@{}lllr@{}}
\toprule
Study & Condition & Metric & Mean \\
\midrule
Temporal & Ordered sequence & NLL & 4.9118 \\
Temporal & Rate-grid control & NLL & 6.9293 \\
Temporal & Shuffled order & NLL & 11.0357 \\
Future statistics & Learned encoder & Loss & 0.1103 \\
Future statistics & Random encoder & Loss & 0.1397 \\
Factor association & True pairing & Accuracy (\%) & 46.6667 \\
Factor association & Shuffled pairing & Accuracy (\%) & 6.6667 \\
\bottomrule
\end{tabular}
\end{table}

\section{Related work}
\paragraph{World models and predictive state.}
World models learn representations that support forecasting and control
\citep{ha2018worldmodels,hafner2019planet,hafner2020dreamer}.
Predictive state representations likewise describe a system through predictions
of future observations \citep{littman2002predictive}.
SpikeWorld uses prediction in two phases: offline to train sensory and action
features, and online to obtain outcome-based correction signals. It stores those corrections outside the trained core.

\paragraph{Spiking temporal representations.}
Surrogate gradients enable optimization through discrete spike thresholds
\citep{neftci2019surrogate,zenke2018superspike}, and can support learning from
spike timing \citep{yu2026beyondrate}. Spiking Transformers combine spike-based
processing with attention \citep{zhou2023spikformer}. Event-based sensing
provides temporal contrast streams \citep{gallego2022event}; our experiments
also use encoded feature sequences from conventional data sources. We study how the shared core predicts these sequences and specify each
stream's input encoding and target.

\paragraph{Online adaptation and compact memory.}
Test-time training uses self-supervised deployment signals to update a model
\citep{sun2020ttt}. Fast weights and differentiable plasticity separate rapidly
changing information from slower learned parameters
\citep{ba2016fastweights,miconi2018plasticity}. SpikeWorld implements a related
separation through a fixed-size residual interface and a separate candidate
score state. The model's original semantic computation remains outside these
online writes.

\paragraph{Adaptive control.}
Online system identification and multiple-model adaptive control estimate
changes and adjust commands \citep{narendra1997multiple,yu2017uposi}.
Learned motor adaptation can infer context across changing robot dynamics
\citep{kumar2021rma}. SpikeWorld's candidate-score path belongs to this
identification-and-correction setting. Its residual matrices add a distinct
learned prediction correction, whose contribution is isolated experimentally.
The appendix compares corrections applied in actuator space and at the
prediction output.

\section{Discussion and conclusion}
We asked where a world model should store what it learns after deployment.
SpikeWorld uses the difference between predicted and observed outcomes to
learn compact external corrections. The trained predictive--semantic model
remains fixed, so its semantic outputs on the same inputs remain unchanged.
This preservation follows from the computation itself and requires no
balancing of online semantic and adaptation losses.

Residual matrices refine predictions; candidate scores select a registered
command correction. Removing each component in turn identifies its effect,
and the closed-loop study shows that action correction improves reward.
These results suggest that a deployment memory should be described by both
the evidence it stores and the computation that reads it. Components trained
from the same observed error can affect prediction and action differently.

The measured configuration improves prediction with a small mutable state
while preserving the shared model. Its temporal core supplies predictive and
semantic features; its external state stores corrections learned from new
outcomes. This separation makes the location and effect of online learning
explicit.

\section*{AI use statement}
Generative AI tools assisted with literature lookup, research discussion,
manuscript drafting and language editing, figure preparation, and preparation
of verification and release code. The authors reviewed the AI-assisted work
and retain responsibility for the scientific claims, experimental records,
and accompanying artifacts. Reported experimental quantities come from the
recorded experiments; this revision introduces no new training runs.

\section*{Reproducibility statement}
Section~\ref{sec:model} defines the model and training objectives;
Section~\ref{sec:adapt} defines the online interface.
Appendix~\ref{app:training} details the data encodings and training settings,
Appendix~\ref{app:algorithm} specifies the update schedule and memory ledger,
and Appendix~\ref{app:extended} gives experimental units, statistical methods,
and additional comparisons. Each comparison identifies which observations,
parameters, and adaptation states are shared across methods.
The public repository at \url{https://github.com/Oooorca/SpikeWorld}
provides reference code, recorded experimental outputs, and scripts for
checking the reported results. Fresh training and simulation runs require
the corresponding datasets and model checkpoints.

\begingroup
\bibliographystyle{iclr2027_conference}
\bibliography{references}

\begin{thebibliography}{17}
\providecommand{\natexlab}[1]{#1}
\providecommand{\url}[1]{\texttt{#1}}
\expandafter\ifx\csname urlstyle\endcsname\relax
  \providecommand{\doi}[1]{doi: #1}\else
  \providecommand{\doi}{doi: \begingroup \urlstyle{rm}\Url}\fi

\bibitem[Ba et~al.(2016)Ba, Hinton, Mnih, Leibo, and
  Ionescu]{ba2016fastweights}
Jimmy Ba, Geoffrey~E. Hinton, Volodymyr Mnih, Joel~Z. Leibo, and Catalin
  Ionescu.
\newblock Using fast weights to attend to the recent past.
\newblock In \emph{Advances in Neural Information Processing Systems},
  volume~29, 2016.
\newblock URL
  \url{https://proceedings.neurips.cc/paper/2016/hash/9f44e956e3a2b7b5598c625fcc802c36-Abstract.html}.

\bibitem[Cramer et~al.(2022)Cramer, Stradmann, Schemmel, and
  Zenke]{cramer2022heidelberg}
Benjamin Cramer, Yannik Stradmann, Johannes Schemmel, and Friedemann Zenke.
\newblock The heidelberg spiking data sets for the systematic evaluation of
  spiking neural networks.
\newblock \emph{IEEE Transactions on Neural Networks and Learning Systems},
  33\penalty0 (7):\penalty0 2744--2757, 2022.
\newblock \doi{10.1109/TNNLS.2020.3044364}.

\bibitem[Gallego et~al.(2022)Gallego, Delbruck, Orchard, Bartolozzi, Taba,
  Censi, Leutenegger, Davison, Conradt, Daniilidis, and
  Scaramuzza]{gallego2022event}
Guillermo Gallego, Tobi Delbruck, Garrick Orchard, Chiara Bartolozzi, Brian
  Taba, Andrea Censi, Stefan Leutenegger, Andrew~J. Davison, Jorg Conradt,
  Kostas Daniilidis, and Davide Scaramuzza.
\newblock Event-based vision: A survey.
\newblock \emph{IEEE Transactions on Pattern Analysis and Machine
  Intelligence}, 44\penalty0 (1):\penalty0 154--180, 2022.
\newblock \doi{10.1109/TPAMI.2020.3008413}.

\bibitem[Ha \& Schmidhuber(2018)Ha and Schmidhuber]{ha2018worldmodels}
David Ha and J{\"u}rgen Schmidhuber.
\newblock Recurrent world models facilitate policy evolution.
\newblock In \emph{Advances in Neural Information Processing Systems},
  volume~31, 2018.
\newblock URL
  \url{https://proceedings.neurips.cc/paper/2018/hash/2de5d16682c3c35007e4e92982f1a2ba-Abstract.html}.

\bibitem[Hafner et~al.(2019)Hafner, Lillicrap, Fischer, Villegas, Ha, Lee, and
  Davidson]{hafner2019planet}
Danijar Hafner, Timothy Lillicrap, Ian Fischer, Ruben Villegas, David Ha,
  Honglak Lee, and James Davidson.
\newblock Learning latent dynamics for planning from pixels.
\newblock In \emph{Proceedings of the 36th International Conference on Machine
  Learning}, volume~97 of \emph{Proceedings of Machine Learning Research}, pp.\
   2555--2565. PMLR, 2019.
\newblock URL \url{https://proceedings.mlr.press/v97/hafner19a.html}.

\bibitem[Hafner et~al.(2020)Hafner, Lillicrap, Ba, and
  Norouzi]{hafner2020dreamer}
Danijar Hafner, Timothy Lillicrap, Jimmy Ba, and Mohammad Norouzi.
\newblock Dream to control: Learning behaviors by latent imagination.
\newblock In \emph{International Conference on Learning Representations}, 2020.
\newblock URL \url{https://openreview.net/forum?id=S1lOTC4tDS}.

\bibitem[Kumar et~al.(2021)Kumar, Fu, Pathak, and Malik]{kumar2021rma}
Ashish Kumar, Zipeng Fu, Deepak Pathak, and Jitendra Malik.
\newblock {RMA}: Rapid motor adaptation for legged robots.
\newblock In \emph{Robotics: Science and Systems XVII}, 2021.
\newblock \doi{10.15607/RSS.2021.XVII.011}.

\bibitem[Littman et~al.(2001)Littman, Sutton, and Singh]{littman2002predictive}
Michael~L. Littman, Richard~S. Sutton, and Satinder Singh.
\newblock Predictive representations of state.
\newblock In \emph{Advances in Neural Information Processing Systems},
  volume~14, pp.\  1555--1561, 2001.

\bibitem[Miconi et~al.(2018)Miconi, Stanley, and Clune]{miconi2018plasticity}
Thomas Miconi, Kenneth~O. Stanley, and Jeff Clune.
\newblock Differentiable plasticity: Training plastic neural networks with
  backpropagation.
\newblock In \emph{Proceedings of the 35th International Conference on Machine
  Learning}, volume~80 of \emph{Proceedings of Machine Learning Research}, pp.\
   3559--3568. PMLR, 2018.
\newblock URL \url{https://proceedings.mlr.press/v80/miconi18a.html}.

\bibitem[Narendra \& Balakrishnan(1997)Narendra and
  Balakrishnan]{narendra1997multiple}
Kumpati~S. Narendra and Jeyendran Balakrishnan.
\newblock Adaptive control using multiple models.
\newblock \emph{IEEE Transactions on Automatic Control}, 42\penalty0
  (2):\penalty0 171--187, 1997.
\newblock \doi{10.1109/9.554398}.

\bibitem[Neftci et~al.(2019)Neftci, Mostafa, and Zenke]{neftci2019surrogate}
Emre~O. Neftci, Hesham Mostafa, and Friedemann Zenke.
\newblock Surrogate gradient learning in spiking neural networks: Bringing the
  power of gradient-based optimization to spiking neural networks.
\newblock \emph{IEEE Signal Processing Magazine}, 36\penalty0 (6):\penalty0
  51--63, 2019.
\newblock \doi{10.1109/MSP.2019.2931595}.

\bibitem[Sun et~al.(2020)Sun, Wang, Liu, Miller, Efros, and Hardt]{sun2020ttt}
Yu~Sun, Xiaolong Wang, Zhuang Liu, John Miller, Alexei~A. Efros, and Moritz
  Hardt.
\newblock Test-time training with self-supervision for generalization under
  distribution shifts.
\newblock In \emph{Proceedings of the 37th International Conference on Machine
  Learning}, volume 119 of \emph{Proceedings of Machine Learning Research},
  pp.\  9229--9248. PMLR, 2020.
\newblock URL \url{https://proceedings.mlr.press/v119/sun20b.html}.

\bibitem[Yu et~al.(2020)Yu, Quillen, He, Julian, Hausman, Finn, and
  Levine]{yu2019metaworld}
Tianhe Yu, Deirdre Quillen, Zhanpeng He, Ryan Julian, Karol Hausman, Chelsea
  Finn, and Sergey Levine.
\newblock {Meta-World}: A benchmark and evaluation for multi-task and meta
  reinforcement learning.
\newblock In \emph{Proceedings of the Conference on Robot Learning}, volume 100
  of \emph{Proceedings of Machine Learning Research}, pp.\  1094--1100. PMLR,
  2020.
\newblock URL \url{https://proceedings.mlr.press/v100/yu20a.html}.

\bibitem[Yu et~al.(2017)Yu, Tan, Liu, and Turk]{yu2017uposi}
Wenhao Yu, Jie Tan, C.~Karen Liu, and Greg Turk.
\newblock Preparing for the unknown: Learning a universal policy with online
  system identification.
\newblock In \emph{Robotics: Science and Systems XIII}, 2017.
\newblock \doi{10.15607/RSS.2017.XIII.048}.

\bibitem[Yu et~al.(2026)Yu, Sun, and Goodman]{yu2026beyondrate}
Ziqiao Yu, Pengfei Sun, and Dan F.~M. Goodman.
\newblock Beyond rate coding: Surrogate gradients enable spike timing learning
  in spiking neural networks.
\newblock \emph{Neuromorphic Computing and Engineering}, 6:\penalty0 014016,
  2026.
\newblock \doi{10.1088/2634-4386/ae46d5}.

\bibitem[Zenke \& Ganguli(2018)Zenke and Ganguli]{zenke2018superspike}
Friedemann Zenke and Surya Ganguli.
\newblock {SuperSpike}: Supervised learning in multilayer spiking neural
  networks.
\newblock \emph{Neural Computation}, 30\penalty0 (6):\penalty0 1514--1541,
  2018.
\newblock \doi{10.1162/NECO_A_01086}.

\bibitem[Zhou et~al.(2023)Zhou, Zhu, He, Wang, Yan, Tian, and
  Yuan]{zhou2023spikformer}
Zhaokun Zhou, Yuesheng Zhu, Chao He, Yaowei Wang, Shuicheng Yan, Yonghong Tian,
  and Li~Yuan.
\newblock Spikformer: When spiking neural network meets transformer.
\newblock In \emph{International Conference on Learning Representations}, 2023.
\newblock URL \url{https://openreview.net/forum?id=frE4fUwz_h}.

\end{thebibliography}
\endgroup
\clearpage
\appendix
\section{Protocols and extended experimental evidence}
\label{app:extended}
\subsection{Experimental units and statistics}
\label{app:protocols}
Table~\ref{tab:studies} separates the study units. A foundation is a sensory
checkpoint; an action seed initializes or trains its action path. Multiple
action seeds under one foundation are not treated as independent foundations.
\begin{table}[htbp]\centering\small
\caption{Experimental units and the question answered by each study.}
\label{tab:studies}
\begin{tabular}{@{}p{.24\linewidth}p{.31\linewidth}p{.39\linewidth}@{}}\toprule
Study & Units & Question\\\midrule
Main stream & 3 foundations, 3 action seeds each & Prediction, tracking, storage, return correction, latency\\
Routed actuator RLS & Same $3\times3$ layout, separate replay & Actuator-space identification under the six-stage protocol\\
Component roles & 5 model-seed clusters & Matrix contribution versus score-only and no-inverse variants\\
Behavioral equivalence & 4 controllers, 3 tasks, 3 environment seeds & Bounded versus legacy history; first changed command\\
Closed loop & 5 models, 3 tasks, 5 replicas per arm & Reward and success under attenuation\\
Joint learning & 5 action seeds, one sensory foundation & Shared-core predictive and semantic learning\\
Training-stage pilot & One foundation; matched cached stream & Same adaptation interface across offline stages\\
Controlled probes & 3 seeds per separate model study & Temporal, future-statistic, and factor-association tests\\
Output calibration & One foundation; 4 paired units per arm & Hidden versus output-space correction\\\bottomrule
\end{tabular}\end{table}

For the main stream, each action-seed run first averages the measured shifted
transition errors, then computes its gain relative to the paired frozen arm.
There are 126 measured shifted transitions per run, after omitting each
episode's four-transition prefix from the score. All methods replay the same
observations; tracking is evaluated through the known actuator map.
Action seeds are averaged within each foundation, and the reported mean
averages foundations. The 95\% intervals use 20,000 bootstrap draws
of foundation means. Update latency uses 2,988 synchronized samples per
adaptive arm after discarding 16 warm-up updates from each run.

Source stages reuse router-calibration episodes and four-transition prefixes;
the provenance record identifies 48 reused source-prefix rows per evaluated
replica. Shifted stages use a separate execution archive. This defines the
calibrated-source reentry measurement in the main table. The sequence can
retain matrices and scores across stages while re-establishing the route at
the start of each episode.

Closed-loop statistics preserve method pairing while resampling model, task,
and replica axes independently, with 20,000 draws. Success is the episode-level
indicator of task completion; cumulative reward sums environment rewards.
This hierarchical pairing is retained when estimating reward differences
between methods.

\subsection{Independent foundation results}
\begin{table}[tbp]
\centering\small
\caption{Prediction gain by independent foundation (\%). Each row averages three action seeds.}
\label{tab:foundations}
\begin{tabular}{@{}lrrrr@{}}
\toprule
Foundation & External state & Adapter & Prediction-path & Replay \\
\midrule
33401 & 21.80 & 14.60 & 25.30 & 25.08 \\
33402 & 20.34 & 17.42 & 17.47 & 16.76 \\
33403 & 15.71 & 14.82 & 22.79 & 19.29 \\
\bottomrule
\end{tabular}
\end{table}

The three foundation rows show how prediction gain varies across independently
trained foundations. The frozen reference has zero gain by definition. The main result is a
mean trade-off over these independently trained foundations, with each
foundation shown separately in Table~\ref{tab:foundations}.

The same-protocol routed actuator RLS uses 4,112 bytes, including its routing
prefix. It obtains 0.58\% prediction gain and $-1.05$\% tracking gain; the
replicated external-state arm obtains 19.72\% and 3.87\%, respectively.
Both return zero correction on calibrated source revisits. This RLS estimates
actuator coupling and is distinct from the output-space calibrators below.

\subsection{Controlled temporal and cross-modal probes}
\label{app:probes}
These studies use separate event encoders trained with predictive and
contrastive objectives. They test representation properties independently
of the main deployment checkpoints.
Identity metadata builds paired views; the future-prediction targets themselves
are computed from observations.

In the temporal probe, the full sequence is compared with two controls.
Order shuffling permutes non-time event features across fixed timestamps.
The rate-grid control sorts event features and assigns evenly spaced
timestamps. The outcome is future-slice negative log-likelihood (NLL), where
lower values indicate better prediction. These controls test sensitivity to
the temporal structure of this task.

The future-statistic probe predicts normalized histograms relating event
time and feature coordinates in later image/audio views. A learned encoder
is compared with a random encoder using the same prediction task. The
composition probe receives factor segments and compares true image/audio
factor associations with shuffled associations. Its score is top-1 factor
association accuracy. Table~\ref{tab:probes} reports the three-seed
means; the separate objectives have different metric scales.

\subsection{Three-stage joint-learning results}
\label{app:joint}
Table~\ref{tab:joint_stages} reports all three stages of the five-action-seed
study. Its sensory metric averages four groups: audio (the mean of SHD and
SSC), image, text, and video. Each sensory loss is normalized by its persistence baseline. The reported
aggregate gives equal weight to the four groups, so the two audio sources
together have the same weight as each other modality.
Relative to action-only updates, joint training increases action MSE by
3.30\% on average across paired seeds (95\% CI: 1.65\%--4.94\%), while
improving class retrieval by 3.31 percentage points (95\% CI: 0.94--5.38).
Joint updates improve retrieval, while action-only updates favor action
prediction.
\begin{table}[tbp]
\centering\small
\caption{All stages of the five-action-seed joint-learning study. Sensory MSE is the four-group normalized mean; accuracy and retrieval are percentages.}
\label{tab:joint_stages}
\begin{tabular}{@{}lrrrr@{}}
\toprule
Stage & Action MSE & Sensory MSE & Semantic (\%) & Retrieval (\%) \\
\midrule
Sequential & 0.1463 & 0.6776 & 19.1488 & 10.0000 \\
Action-only shared update & 0.1174 & 0.6786 & 19.2655 & 8.5000 \\
Joint shared update & 0.1212 & 0.6725 & 20.9440 & 11.8125 \\
\bottomrule
\end{tabular}
\end{table}

\subsection{Training-stage comparison}
\label{app:bridge}
The pilot begins with one common sensory-pretrained checkpoint. From the
sequential action-trained model, three copies receive 200 additional updates:
joint training, action loss alone, or joint training without the sensory
next-slice term. Thus the no-next variant removes the term in the final stage,
not from the full pretraining history. Offline fitting supplies each model's
basis and router using the same procedure and hyperparameters.

All models replay the same cached six-stage trajectory and compare frozen,
score-only, and full external state on 126 shifted transitions after routing.
Table~\ref{tab:bridge} shows that the interface can be used across these
training stages. The ordering of adapted errors differs from the ordering of offline action
errors. Lower offline action error alone does not determine which checkpoint
has the lowest error after adaptation in this pilot.
\begin{table}[tbp]
\centering\small
\caption{One-foundation training-stage pilot. MSE is evaluated on 126 shifted transitions after the observation prefixes. All variants inherit sensory predictive pretraining.}
\label{tab:bridge}
\begin{tabular}{@{}lrrrr@{}}
\toprule
Training stage & Frozen MSE & Score MSE & Full MSE & Gain (\%) \\
\midrule
Sequential & 0.3927 & 0.3661 & 0.2878 & 26.72 \\
Action-only & 0.4408 & 0.3797 & 0.2918 & 33.81 \\
Joint & 0.4097 & 0.3992 & 0.3116 & 23.96 \\
Joint, no-next & 0.4270 & 0.4129 & 0.3186 & 25.39 \\
\bottomrule
\end{tabular}
\end{table}

\subsection{Direct output-space calibration}
\label{app:additional}
The additional development comparison applies corrections at the output
rather than the hidden feature. Table~\ref{tab:additional} preserves all
methods: hidden residual state, direct-output NLMS, direct-output RLS, and a
low-rank readout update. It uses one sensory foundation and four paired units
per method, rather than the three-foundation main comparison.
\begin{table}[tbp]
\centering\small
\caption{Separate output-calibration development protocol: one sensory foundation, four paired units per method. These results are not pooled with the main-stream comparison.}
\label{tab:additional}
\begin{tabular}{@{}lrrrr@{}}
\toprule
Correction & MSE & Gain (\%) & Bytes & Update (ms) \\
\midrule
Hidden residual state & 0.2640 & 23.34 & 28,848 & 11.70 \\
Output NLMS & 0.1652 & 51.70 & 13,956 & 11.03 \\
Output RLS & 0.1839 & 45.84 & 21,156 & 11.10 \\
Low-rank readout & 0.2961 & 14.14 & 30,532 & 11.97 \\
\bottomrule
\end{tabular}
\end{table}

In this setting, direct-output NLMS and RLS achieve larger prediction gains
with less mutable state than the hidden-state correction. This comparison favors output-space correction in this development setting.
The protocols differ, so these values are interpreted within this comparison
rather than pooled with the main-stream frontier.

\subsection{Closed-loop detail}
\label{app:control}
Each model/task controller is initialized once, then processes five fresh
trajectory replicas in order. External state resets its route at each episode
but keeps its matrices and candidate scores. Scalar RLS and gain-memory replay
also retain their accumulated state across these replicas. All episodes use
attenuation 0.4 and last 13 decisions, or 104 simulator steps.

The live-control tuning arm uses AdamW at learning rate $10^{-5}$ on parameters
whose names do not contain \texttt{semantic}. Gain-memory replay retrieves
local gain estimates from observation/task features using nearest neighbors,
with capacity 141. It does not train the prediction path with replay as the
main-stream Adam arm does. The scalar RLS controller accumulates a numerator
and denominator using finite-difference model sensitivity. The oracle divides
by the true attenuation and clips to the valid command range; unlike the
external-state controller, it has no 0.05 correction cap. Consequently, the
main-stream storage and latency entries are not assigned to these live-control
configurations.

All six arms appear in Table~\ref{tab:control}, including the scalar RLS and
oracle references. The external-state reward gains by task are 2.31 for
pick-place, 11.61 for push, and 9.79 for reach. The five model-seed gains range
from 7.19 to 8.66. Prediction-model tuning and external state both achieve
66.67\% aggregate success. The scalar RLS obtains the largest reward among
the non-oracle methods in this single-fault control protocol. External state improves reward over frozen control, while scalar RLS is the
stronger controller for this particular fault.

\section{Data, architecture, and training}
\label{app:training}
This appendix defines the input encodings, architecture, and training objectives.
Appendix~\ref{app:notation} collects the symbols. The sensory sequence
clock is indexed by $\tau$; the robot's macro-decision clock is indexed by $n$.
One macro-decision starts at environment step $t_n$ and has nominal horizon
$H=8$. In the closed-loop runner, the transformed command is held for that
interval, unless the episode terminates or reaches its remaining-step limit.

\subsection{Data interfaces and the shared core}
The audio streams use 224 channels for SHD and 285 for SSC. SHD input is
zero-padded to the shared audio stem's width, while the corresponding output
heads retain their original dimensions. Retinal image and rendered-word
streams each use 40 channels and share an input stem. The video stream uses
896 pooled feature channels. All sensory sequences have length 16. Retinal
data use the handwritten-digit dataset distributed with scikit-learn;
rendered words are represented by spatial features under a common scanpath.
Video inputs are causal feature changes from a content-disjoint
Something-Something-V2 cache. The study uses cached inputs with their recorded
splits rather than collecting physical event-camera measurements.

The model has width 96, two shared blocks, four attention heads, and a
feed-forward width of 192. Input stems apply layer normalization and linear
projection, followed by position and modality embeddings. Input spikes are
added to the projected numerical features. Each shared block has spike-based
attention and a spiking feed-forward path, both with residual connections.
The output is layer-normalized. Sensory next-slice prediction adds the learned
output-head change to the current slice, so copying the current feature is
the natural zero-change reference.

The attention implementation restricts each query to a local causal window
of at most four positions, rather than selecting the four largest scores
from all past positions. Queries, keys, and values are formed with LIF units;
membrane-dependent gates modulate attention scores, which are normalized by
softmax within the available window. This limits the number of attention pairs. We do not measure neuromorphic
energy in these experiments. LIF state is initialized
for each input sequence; deployment memory resides in the explicit external
state rather than an unbounded cross-episode membrane history.

The action branch embeds the 39-coordinate observation and four-coordinate
command into two tokens. Task and token-type embeddings are added to them,
and the same core processes the pair. Its final token provides the action
feature $h_n$ for the linear readout. The complete checkpoint has 1,451,388
parameters. The joint-training configuration exposes 605,079 trainable
parameters; the main deployment interface freezes the selected checkpoint.

\subsection{Full offline objectives}
\label{app:losses}
Let $\ell_{\rm SHD}$ and $\ell_{\rm SSC}$ be Equation~\eqref{eq:next} on the
two audio sources, and let $\ell_{\rm image},\ell_{\rm text},\ell_{\rm video}$
be the other sensory losses. The grouping used in training is
\begin{equation}
 \mathcal L_{\mathrm{next}}=\frac14\left[
 \frac{\ell_{\rm SHD}+\ell_{\rm SSC}}2+
 \ell_{\rm image}+\ell_{\rm text}+\ell_{\rm video}\right].
\end{equation}
Audio receives one group weight despite having two sources. The persistence
denominator is floored at $10^{-8}$ to prevent division by zero.
This equation defines the training loss, whose denominators come from the
training split. Evaluation aggregation is specified separately in
Appendix~\ref{app:joint}.

For a batch of $N$ labeled examples, let $y_i$ be the class of example $i$
and let $p_{i,y_i}$ be its predicted softmax probability. Semantic cross-entropy
is $-N^{-1}\sum_{i=1}^N\log p_{i,y_i}$, averaged across the five sensory
streams. The coupled heads flatten the length-16 hidden sequence, normalize
it, apply dropout during training, and predict the relevant classes. The
class counts are 20 for SHD, 35 for SSC, 10 for retinal digits/words, and 42
for the video cache. An inherited stable semantic branch separately operates
on standardized flattened sensory inputs. Both branches are frozen at
deployment; the joint-learning semantic results concern the coupled branch.

For binding, $N$ is the number of paired image/text examples in the batch.
The subscripts $I$ and $T$ below denote image and text. Let $u_i$ and $v_j$ be
unit-normalized 64-dimensional image and
text representations, respectively. These are distinct from the membrane
symbol $u_\tau$ in Equation~\eqref{eq:lif}. Define similarity
$b_{ij}=u_i^\top v_j/\tau_b$, where the temperature is $\tau_b=0.12$.
Let $P_i$ contain all text indices whose digit class matches image $i$.
The image-to-text loss is
\begin{equation}
 \mathcal L_{I\rightarrow T}=-\frac1N\sum_{i=1}^N
 \log\frac{\sum_{j\in P_i}\exp(b_{ij})}{\sum_{j=1}^N\exp(b_{ij})}.
\end{equation}
This assigns probability mass to all matching-class texts, rather than one
arbitrarily chosen pair. The reverse loss $\mathcal L_{T\rightarrow I}$
exchanges image and text roles, and
$\mathcal L_{\mathrm{bind}}=(\mathcal L_{I\rightarrow T}+\mathcal L_{T\rightarrow I})/2$.
The binding representation averages hidden features across sequence positions
before its projection.

The action objective is Equation~\eqref{eq:action}. Its mean and scale are
fit from offline observation changes and reused during deployment. The
weights in Equation~\eqref{eq:joint} are the fixed configuration used by the
recorded study. Sequential action training runs for 800 updates. The additional
joint stage uses 200 updates, batch size 32, and learning rate $10^{-4}$.
Sensory prediction and action prediction both supply gradients to the shared
core in this joint stage.

\section{Online algorithm and state accounting}
\label{app:algorithm}
\subsection{Initialization, routing, and state lifetime}
The two residual matrices and all candidate scores start at zero. The stored
counter starts at zero. At the start of each episode, the current route is
source, the route is unlocked, and the temporary signature and pending-loss
lists are cleared. The matrices and cumulative scores persist across episodes
within a run. A new run creates a new state.

The four routing classes are source, shear, attenuation, and noise. The router
is a frozen two-layer network with a 64-unit ReLU hidden layer, fit offline
using calibration signatures. Online, it receives the mean signature of the
first four transitions. A signature contains candidate losses relative to identity, signed and
absolute identity prediction errors, signed and absolute commands, and current
physical observations. It also contains signed and absolute observation changes
and cross-products between the first two command coordinates and the first
six observation-change coordinates. These concatenate to 247
values per transition. The true mechanism label is used to fit the router
offline but is absent from its online input.

Once the fourth signature arrives, the router locks for that episode. For
shear or attenuation, the four pending loss vectors are added to the persistent
scores, and their count is committed. Subsequent outcomes add one new loss
vector each. Source and noise routes keep the identity command transformation
and do not update residual matrices. Both temporary lists are cleared when
the route locks. The scores are shared across tasks and indexed by candidate;
the two matrices are indexed by mechanism family.

The selected candidate is the minimum cumulative score within the locked
family. Dividing scores by the shared count would leave this ordering
unchanged. Ties use the first minimum in the fixed bank order.
Scores are computed from the frozen
predictor, before the residual-matrix correction.

\subsection{Candidate transformations and command correction}
The identity candidate is followed by 15 shear values
$k\in\{0.05,0.10,\ldots,0.75\}$ and 16 attenuation values
$g\in\{0.20,0.25,\ldots,0.95\}$. For command coordinates $(a_x,a_y)$,
\begin{align}
 T_{\rm shear}(a;k)_x&=a_x,&
 T_{\rm shear}(a;k)_y&=a_y+ka_x,\\
 T_{\rm attenuation}(a;g)_{x,y}&=g\,a_{x,y}.&&
\end{align}
Other coordinates are unchanged. Transformed candidates are clipped to
$[-1,1]$ before querying the predictor. The inferred inverse uses
$\tilde a_y=a_y^*-\hat k a_x^*$ for shear, or
$\tilde a_{x,y}=a_{x,y}^*/\max(\hat g,0.20)$ for attenuation.
The hat denotes the selected bank value. The floor $0.20$ prevents excessive
division and coincides with the lowest attenuation candidate. The change
from the desired command is then capped by Equation~\eqref{eq:command}.
If the resulting command exceeds the action bounds, the entire proposal is
replaced by the desired command for that example.

\paragraph{Worked command example.}
For a one-coordinate illustration, suppose the route has locked and prior
observations have selected gain 0.5. A desired command of 0.10 gives inverse
proposal 0.20. The 0.05 cap makes the executed command 0.15. If the actual gain
is 0.5, its realized effect is 0.075, compared with 0.05 without correction.
Candidate scoring uses the forward gain transform to explain observed outcomes;
the inverse compensates a later command. The residual matrix below changes the
predicted outcome, not this command. These illustrative values explain the
operation rather than report an additional experiment.

\subsection{Exact matrix write}
\label{app:matrix-update}
The fixed basis $B$ contains the first 30 right singular vectors of centered
offline action hidden features. After observing $d_n$ and committing the
candidate-score update, the prediction used for the matrix write is recomputed
with the selected candidate and the current matrix. Let $C$ denote this
active matrix, and let
\begin{equation}
 L_n(C)=\frac1{D_q}\|W_{\rm out}(h_n+Cz_n)+b_{\rm out}-d_n\|_2^2,
 \qquad A_n=\nabla_C L_n(C).
\end{equation}
Here $L_n$ is the current matrix-write loss, $A_n$ is its matrix gradient, and
all other quantities are held fixed
for this write. The normalized step and projection are
\begin{equation}
 C'=C-\eta\frac{A_n}{\max(\|A_n\|_F,\epsilon)},\qquad
 C\leftarrow C'\min\!\left(1,\frac{R_C}{\max(\|C'\|_F,\epsilon)}\right),
 \quad \epsilon=10^{-12}.
 \label{eq:update}
\end{equation}
$C'$ is the tentative matrix before projection, and $\epsilon$ safeguards
division by a zero norm. The Frobenius norm $\|\cdot\|_F$ is the square root of the sum of squared
matrix entries. Normalization controls step length; projection bounds the
matrix norm. A zero gradient yields a zero step. No optimizer moments are
maintained. This recomputed training loss is distinct from the pre-update
prediction error already recorded for evaluation.

\paragraph{A useful interpretation.}
Let $e_n=d_n-g_{\theta_0}(q_n,T_j(a_n),c)$ be the uncorrected error after
candidate selection, and let $v_n=W_{\rm out}C_\rho z_n$ be the correction in
output space. Expanding the square gives
\begin{equation}
 \frac{\|e_n\|_2^2-\|e_n-v_n\|_2^2}{D_q}
 =\frac{2e_n^\top v_n-\|v_n\|_2^2}{D_q}.
\end{equation}
The first term rewards alignment with the residual; the second penalizes an
oversized correction. Thus prediction improvement depends on its direction
as well as its magnitude. This identity concerns prediction error, while
the separate inverse determines command changes.

\subsection{One decision cycle}
\begin{enumerate}[leftmargin=*,itemsep=3pt]
\item Read the current physical observation and desired policy command.
      Use the previously available route and scores to form a capped command.
\item Predict its outcome with the frozen core and the currently available
      residual matrix. Save this prediction for evaluation.
\item Execute the command and wait for the outcome. In live attenuation
      control, the actuator-transformed command is held over the macro horizon.
\item Score the saved prediction against the observed outcome before writing
      either external component.
\item Compute all candidate losses with the uncorrected frozen predictor.
      Before route lock, append the bounded signature/loss prefix. At the
      fourth outcome, infer the family and, for shear or attenuation, commit
      the four pending loss vectors exactly once. Clear the temporary prefix.
      On subsequent outcomes in these routes, add only the new loss vector.
\item If the route is now locked to shear or attenuation, perform one matrix
      write with Equation~\eqref{eq:update}. The fourth outcome is the first
      eligible write; its candidate losses are not added a second time.
\item Continue with the next decision. The first opportunity for a changed
      command is decision five.
\end{enumerate}
The ordering keeps evaluation before learning: an observed outcome can
change subsequent predictions and commands, but never its already-scored
prediction.

\subsection{Complete mutable-state ledger}
\begin{table}[tbp]\centering\small
\caption{Peak mutable numeric allocation in FP32. Fixed model/router/basis
parameters and transient activations are outside this adaptation-state metric.}
\label{tab:ledger}
\begin{tabular}{@{}llr@{}}\toprule
Object & Arithmetic / lifetime & Bytes\\\midrule
Two residual matrices & $2\times96\times30\times4$; stream & 23,040\\
Candidate scores and count & $(32+1)\times4$; stream & 132\\
Compatibility padding & $303\times4$; fixed allocation & 1,212\\
Routing signatures and losses & $4\times(247+32)\times4$; prefix & 4,464\\\midrule
Peak total & Tensor block plus prefix & 28,848\\\bottomrule
\end{tabular}\end{table}
The allocated tensor block is 24,384 bytes. The active entries use 23,172
bytes; the remaining entries preserve the historical allocation layout.
The prefix is bounded at four observations and released on route lock.
Python-object overhead and transient inference/gradient buffers are not
included in this numeric-state ledger. The same persistent-state accounting
includes trainable prediction parameters, optimizer tensors, and replay for
the comparison arms. Bytes are not a hardware energy measurement.

\section{Notation at a glance}
\label{app:notation}
\begin{longtable}{@{}p{.21\linewidth}p{.27\linewidth}p{.46\linewidth}@{}}
\caption{Principal symbols. Scalars, dimensions, learned quantities, and fixed
offline quantities are identified separately.}\label{tab:notation}\\
\toprule Symbol & Shape or value & Meaning \\\midrule\endfirsthead
\toprule Symbol & Shape or value & Meaning \\\midrule\endhead
$m,\tau,T,D_m$ & Stream; position; $16$; stream-specific & Sensory stream, sequence index, length, and channel count.\\
$x^m_\tau,\hat x^m_{\tau+1}$ & $\R^{D_m}$ & Observed current and predicted next feature slice.\\
$p_m$ & Positive scalar & Persistence MSE used to normalize the sensory objective.\\
$I_\tau,v_\tau,u_\tau$ & Per-neuron values & Input current, pre-reset membrane, and post-reset membrane.\\
$s_\tau,\beta,V_{\rm th}$ & Binary; $0.85$; $1$ & Spike, decay, and firing threshold.\\
$n,t_n,H$ & Decision; environment step; $8$ & Separate macro-decision and simulator clocks.\\
$q_n,q_n^+,D_q$ & $\R^{39}$; $39$ & Current/outcome physical observations and their dimension.\\
$a_n^*,\tilde a_n,a_n$ & $[-1,1]^4$ for valid commands & Desired, inverse-proposed, and executed commands.\\
$c$ & Task identifier & One of the four manipulation task embeddings.\\
$d_n,\hat d_n$ & $\R^{39}$ & Standardized observed and predicted observation change.\\
$\mu_\Delta,\sigma_\Delta$ & $\R^{39}$, fixed & Offline mean and positive scale of observation changes.\\
$\theta,\theta_0,g_\theta$ & Trained parameters / predictor & Offline parameters; frozen deployment copy; endpoint predictor.\\
$h_n,\tilde h_n,d$ & $\R^{96}$; $96$ & Original/corrected action feature and hidden width.\\
$\bar h,B,r,z_n$ & $\R^{96}$; $\R^{30\times96}$; $30$; $\R^{30}$ & Fixed mean, fixed basis, basis rank, and compressed feature.\\
$\rho,C_\rho$ & Route; $\R^{96\times30}$ & Inferred mechanism and its online residual matrix.\\
$W_{\rm out},b_{\rm out}$ & $\R^{39\times96}$; $\R^{39}$ & Frozen linear endpoint readout.\\
$j,J,T_j,S_j$ & Index; $32$; map; scalar & Candidate, bank size, command transform, and cumulative loss.\\
$k,g,\hat k,\hat g$ & Scalars & Shear/gain and their selected candidate estimates.\\
$\eta,R_C,\delta$ & $0.02$, $0.20$, $0.05$ & Matrix step length, matrix norm bound, command correction cap.\\
$\lambda_b,\lambda_a$ & $0.25$, $4$ & Fixed binding/action loss weights in joint training.\\
$E_M,G_M$ & MSE; percentage & Error of method $M$ and gain relative to frozen prediction.\\
$\ell_m,\mathcal L_{\rm next}$ & Scalar losses & Per-stream and grouped sensory prediction objectives.\\
$\mathcal L_{\rm sem},\mathcal L_{\rm bind},\mathcal L_{\rm act}$ & Scalar losses & Semantic classification, image/text association, and action prediction objectives.\\
$\ell_{n,j}$ & Scalar & Observed transition loss for candidate $j$.\\
$L_n,A_n,C'$ & Scalar; $\R^{96\times30}$; same & Matrix-write loss, its gradient, and tentative matrix before projection.\\
$\epsilon$ & $10^{-12}$ & Safeguard for norm division in the matrix update.\\
$e_n,v_n$ & $\R^{39}$ & Uncorrected prediction residual and output-space correction.\\
$N,y_i,p_{i,y_i}$ & Count; class; probability & Batch size, class label and its predicted probability.\\
$u_i,v_j$ & $\R^{64}$, unit norm & Image and text binding vectors (distinct from membrane variables).\\
$P_i,b_{ij},\tau_b$ & Index set; scalar; $0.12$ & Matching-class texts, image/text similarity and binding temperature.\\
\bottomrule
\end{longtable}

\end{document}